\documentclass[pdflatex,sn-mathphys-num]{sn-jnl}% Math and Physical Sciences Numbered Reference Style
\usepackage{graphicx}%
\usepackage{multirow}%
\usepackage{amsmath,amssymb,amsfonts}%
\usepackage{amsthm}%
\usepackage{mathrsfs}%
\usepackage[title]{appendix}%
\usepackage{xcolor}%
\usepackage{textcomp}%
\usepackage{manyfoot}%
\usepackage{booktabs}%
\usepackage{algorithm}
\usepackage{algorithmicx}
\usepackage{algpseudocode}
\usepackage{listings}%
\usepackage{tabularx}  % 自动调整列宽
\usepackage{ragged2e}  % 更好的对齐控制

\theoremstyle{thmstyleone}%
\theoremstyle{thmstyletwo}%

\theoremstyle{thmstylethree}%

\begin{document}
	\title[Article Title]{Optimizing Energy-based Neural Network Training with Coherent Ising Machine}
	
	%%=============================================================%%
	%% GivenName	-> \fnm{Joergen W.}
	%% Particle	-> \spfx{van der} -> surname prefix
	%% FamilyName	-> \sur{Ploeg}
	%% Suffix	-> \sfx{IV}
	%% \author*[1,2]{\fnm{Joergen W.} \spfx{van der} \sur{Ploeg} 
		%%  \sfx{IV}}\email{iauthor@gmail.com}
	%%=============================================================%%
	
	\author[1]{\fnm{Chen-Rui} \sur{Fan}}%\email{iauthor@gmail.com}
	%\equalcont{These authors contributed equally to this work.}
	\author*[2]{\fnm{Bo} \sur{Lu}}\email{lubo@mail.bnu.edu.cn}
	%\equalcont{These authors contributed equally to this work.}
	
	\author[3]{Zhi-Hong Zhang}
	
	\author[3]{Run-Qing Zhang}
	
	\author[3]{Jing-Wei Wen}
	
	\author[4]{Tie-Jun Wang}
	
	\author*[1]{\fnm{Chuan} \sur{Wang}}\email{wangchuan@bnu.edu.cn}
	
	\affil*[1]{\orgdiv{School of Artificial Intelligence}, \orgname{Beijing Normal University}, \orgaddress{ \city{Beijing}, \postcode{100875}, \country{China}}}
	
	\affil[2]{\orgdiv{Laboratory for Advanced Computing and Intelligence Engineering}, \orgname{Information Engineering University}, \orgaddress{ \city{Zhengzhou}, \postcode{450001}, \country{China}}}
	
	\affil[3]{\orgname{China Mobile (Suzhou) Software Technology Company Limited}, \orgaddress{ \city{Suzhou},  \postcode{215163},  \country{China}}}
	
	\affil[4]{\orgname{School of Science, Beijing University of Posts and Telecommunications}, \orgaddress{ \city{Beijing}, \postcode{100875}, \country{China}}}

	%%==================================%%
	%% Sample for unstructured abstract %%
	%%==================================%%
	
	\abstract{While Ising machines serve as advanced physical solvers for the Ising model, enabling applications in combinatorial optimization and neural network training, their scalability for large-scale neural networks remains constrained by hardware connectivity limitations and suboptimal training methodologies. In this work, we leverage a Coherent Ising Machine (CIM) to train an energy-based neural network using Equilibrium Propagation, achieving performance comparable to existing software-based implementations. We further enhance the algorithm by integrating the Adam optimizer to solve for the ground state of a Hopfield energy network, significantly improving convergence speed and solution accuracy. Additionally, we demonstrate the scalability of our approach across deeper network architectures and convolutional operations. Our results highlight the potential of CIM dynamics as a scalable platform for training complex neural networks, offering a pathway toward energy-efficient implementations via analog circuits, optoelectronics, or integrated photonics. This work establishes a novel physical framework for next-generation AI hardware development.}

	
	%%\pacs[JEL Classification]{D8, H51}
	
	%%\pacs[MSC Classification]{35A01, 65L10, 65L12, 65L20, 65L70}
	
	\maketitle
	
	\section{INTRODUCTION}\label{sec1}
	
	Artificial intelligence (AI) has undergone rapid and transformative advancements, with its applications now permeating diverse fields, such as large language models, computer vision, autonomous vehicle control, and biomedicine. The advent of advanced large-scale neural network architectures, such as Transformer \cite{vaswani2017attention}, Convolutional Neural Networks (CNN)\cite{lecun1998gradient}, and  ResNet \cite{he2016deep}, has brought strong impetus to its development. 
	Neural networks, inspired by the functioning of human brain neurons, update inter-neuronal parameters through Back Propagation(BP), enabling the learning of intricate relationships within data. This capability to process complex datasets positions neural networks as a fundamental tool in the advancement of AI research. However, training large-scale networks demands resource-intensive high-performance computing systems, which are encumbered by challenges such as excessive energy consumption and prolonged training times \cite{xu2018scaling}. For instance, training GPT-3 is estimated to consume nearly 1300 MWh \cite{patterson2021carbon}. This has prompted the exploration of novel physical computing architectures for neural network training, including optical neural networks \cite{lin2018all, xu202111, feldmann2021parallel}, physical neural networks \cite{wright2022deep}, quantum annealing \cite{adachi2015application, hu2019quantum, willsch2020support}, and quantum computing \cite{shi2022parameterized,beer2020training}, which aim to achieve faster training speeds and more efficient outcomes.
	
	In large-scale neural networks, neuronal dynamics are dictated by weight matrices, where the training process systematically minimizes the loss function—a process that directly parallels the energy minimization of an Ising model as it evolves toward its ground state through spin interactions. This fundamental correspondence enables neural network training to be mathematically mapped to an Ising Hamiltonian and physically implemented via artificial spin networks that replicate emergent collective behaviors in physical systems \cite{hopfield1982neural}. Thus far, Ising machine simulating the evolution of the Ising model has been realized in various systems, including electronic circuits \cite{goto2021high, tatsumura2021scaling}, superconducting circuits \cite{razmkhah2024josephson, ucpinar2024scalable}, nanomagnets \cite{litvinenko2023spinwave, sutton2017intrinsic}, optics \cite{marandi2014network, mcmahon2016fully, inagaki2016coherent, pierangeli2019large, honjo2021100,lu2023recent}, and optoelectronic systems \cite{bohm2019poor, cen2022large}. The Coherent Ising Machine (CIM), an optical computing platform based on degenerate optical parametric oscillators (DOPOs), can simulate arbitrarily connected Ising spin systems through its measurement-feedback architecture \cite{yamamoto2017coherent,mohseni2022ising}. 
	Currently, the largest Ising machine can scale up to $10^6$ spins, surpassing central processing units in the time required to obtain high-quality approximate solutions for maximum independent set (MIS) problems with over $10^4$ nodes \cite{takesue2025finding}. It also demonstrates significant potential in neural network training. 
	Although the CIM has shown promise for neural network training, its performance remains suboptimal due to the lack of an efficient mapping strategy for learning tasks. Current state-of-the-art training approaches for Ising-based networks predominantly depend on ultra-fast statistical sampling methods. However, these implementations, particularly for Boltzmann machines, suffer from prohibitively high computational resource requirements \cite{adachi2015application, benedetti2016estimation} and simplistic network structures \cite{bohm2022noise}. Sparse Ising machine, which relies on Gibbs sampling, exhibits sensitivity to network structure and input dimensions \cite{niazi2024training}. Furthermore, the complex constraint conditions associated with QUBO optimization networks \cite{song2023training} result in suboptimal training performance.
	
	The Equilibrium Propagation (EP) algorithm aligns with both biological and physical principles, using two equilibrium states (free state and nudge state) to update the weight matrix between neurons \cite{scellier2017equilibrium, scellier2019equivalence, laborieux2021scaling}. It performs well in supervised machine learning tasks and has been successfully applied to systems such as nonlinear resistor networks \cite{kendall2020training}, coupled phase oscillators \cite{wang2024training, rageau2025training}, and the quantum annealing machine \cite{laydevant2024training}. EP trains a converged cyclic energy network, which is equivalent to minimizing the network's energy. The physical system described by this energy form is well-suited for implementation in an Ising machine. However, existing EP-based quantum annealers face fundamental scalability limitations due to their fixed chip architectures, preventing their application to large-scale neural networks \cite{laydevant2024training}.
	In contrast, CIM offers distinct advantages for large-scale implementations, as it supports arbitrary connectivity and demonstrates superior computational performance in densely connected graphs \cite{hamerly2019experimental}. By integrating the EP framework with CIM dynamics, we can overcome current methodological constraints and enable efficient training of energy-based networks through improved gradient approximation.
	
	In this work, we present a novel neural network training framework that combines CIM with EP. By incorporating the Adam optimizer into the CIM architecture, we develop an efficient ground-state search algorithm that achieves comparable accuracy to conventional software-based methods on the MNIST dataset.
	We present the advantages of the Adam-CIM algorithm in locating the ground state of the Ising model and provide a detailed illustration of how to apply it to Multi-Layer Perceptron (MLP) training, where the Ising model is evolved towards its ground state through the Adam-CIM optimization process.
	Furthermore, we analyze the evolution of energy distribution in neural networks before and after CIM training, elucidating the phase transition dynamics during the optimization process and systematically comparing how different network architectures influence training outcomes.
	Critically, we demonstrate the scalability of our approach by successfully applying CIM to convolutional neural networks (CNNs), thereby establishing its capability for large-scale neural network training. When integrated with state-of-the-art optoelectronic systems, our platform achieves significant reductions in both training time and power consumption, which shows a crucial advancement for practical deployment.
	These findings highlight CIM's dual role as both a powerful neural network trainer and a photonic quantum computing platform. The unique physical properties of CIM, when synergized with EP, enable applications that transcend conventional combinatorial optimization problems, opening new possibilities for energy-efficient AI hardware development.
	
	\section{RELATED WORK}\label{sec2}
	
	\subsection{Physics heuristic algorithm : CIM}
	%The energy of the Ising model network can be described by:

	%The Ising model describes the phase transition process between its ferromagnetic and antiferromagnetic states. For a system with $N$ spins, where each spin has two states, denoted as $\sigma_i = +1$ and $\sigma_i = -1$. The total energy of the system is represented by the Ising energy, denoted as $E_{ising} = \sum_{i>j} J_{ij}\sigma_i \sigma_j + \sum_i h_i \sigma_i,$. The coupling strength between the $i$-th and $j$-th spins is denoted by $J_{ij}$, and is considered as the input to the COPs. To analyze the stability of its fixed points, the evolution of the amplitude $\mu_i$ of the $i$-th DOPO can be represented as:
	The Ising model describes the phase transition process between its ferromagnetic and antiferromagnetic states. For a system with $N$ spins, where each spin has two states, denoted as $\sigma_i = +1$ and $\sigma_i = -1$. The total energy of the system is represented by the Ising energy, denoted as 
	\begin{equation}
		E_{Ising}=-\frac{1}{2}\sum_{i=1}^{N}\sum_{j=1}^{N}J_{ij}\sigma_{i}\sigma_{j}-\sum_{i=1}^{N}h_{i}\sigma_{i}\label{eqIsing_energy}.
	\end{equation}
	Here, it represents the energy of a network with binary spins $\sigma_{i}$ under the constraint matrix $J_{ij}$ and the bias field $h_{i}$.Researchers found the loss functions of some COPs are similar to Ising energy, thus developing new ideas for solving NP-hard problems based on basic principles of physics. In 2013, Zhe Wang et al. proposed a CIM implemented by DOPOs and performed calculations on all instances of the Max-Cut problem\cite{wang2013coherent}. The CIM relies on the nonlinear dynamic process of DOPOs, and the amplitude $x_i$ of the $i$-th DOPO can be represented as:
	
	\begin{equation}
		\frac{dx_i}{dt} = (p-1)x_i - x_i^3 + \sum_{i=1, i\neq j}^N \epsilon J_{ij}x_j,\label{eqmuzhenfu}
	\end{equation}
	$p$ is the pump intensity, $\epsilon$ is the coupling strength between the DOPO pulses, and the sign of the spin amplitude $\sigma_j = \frac{x_j}{|x_j|}$ is usually used as the spin value.
	
	\subsection{EP algorithm based on Energy Neural Network}
	
	Traditional neural network training relies on backpropagation (BP) and gradient descent to iteratively adjust network weights and biases according to output-target discrepancies. While effective in digital implementations, this approach presents two fundamental limitations for physical systems: it violates biological plausibility by requiring non-local error feedback that contradicts known neurobiological mechanisms, and it creates implementation barriers in optical computing where gradient calculation and propagation are physically challenging. These constraints render standard BP fundamentally incompatible with many physical implementations, severely restricting both the application of optical neural networks and the scalability of neuromorphic computing architectures. To address this issue, Scellier and Bengio proposed the EP algorithm \cite{scellier2017equilibrium} in 2017,  a biologically plausible learning framework grounded in energy-based dynamics. The approach builds upon a Hopfield-type energy function $E$,  and it is a kind of Hopfield energy network:
	\begin{equation}
		E_{network} = \frac{1}{2}\sum_{l} s_{i}^{2} - \sum_{i>j} W_{ij} \rho(s_{i}) \rho(s_{j}) - \sum_{i} b_{i} \rho(s_{i})\label{eqEP_energy},
	\end{equation}
	$s$ represents the states of the neurons, $W_{ij}$ is the synaptic weights, $b_{i}$ deontes the neuron biases, $\rho$ denotes the nonlinear activation function, and $\rho(s_{i})$ represents the firing rate of unit $i$.
	$W$ is symmetric, a property essential for implementing EP with Adam-CIM, as CIM requires the coupling matrix to be symmetric. 
	Similar to most supervised learning, we use the mean square error $C=\frac{1}{2}\|v-s_{end}\|^2$ as the loss function. 
	Considering weak coupling, the potential energy term of the loss function corresponds to the perturbation relative to the target label, with the nudge intensity denoted as $\beta$. The total energy is given by $ F = E + \beta C $, where the neuron state $s$ spontaneously evolves toward a low-energy spin configuration.
	For a network with parameter $\theta$, we label the neuron state \(s_{\theta,v}^\beta\), corresponding to the fixed point under the targets \(v\) and nudge parameter \(\beta\), when the energy \( E(\theta,v,s_{\theta,v}^\beta)\) is in a local minimum, and more details can be found in the Method part.
	
	The EP algorithm operates in two distinct phases while maintaining a consistent neural computation mechanism. The key distinction between these phases lies in the injection of external energy, which is modulated by the loss function \(C\). In numerical simulations, the transition between phases is controlled by the parameter \(\beta\). In the free phase, \(\beta=0\), which means that the network spontaneously evolves to the local minimum energy without external interference and finally approaches the first stable state. Then, let $\beta \neq 0$ and the system evolves again, which constitutes the nudge phase. The external energy exerts a gentle nudge on the system towards the target, thereby achieving supervised learning with labeled data. The problem corresponds to a constrained optimization problem:
	\begin{equation}
		\frac{\partial E}{\partial s} = 0, \quad \arg \min C(v, s)\label{eqconstrained}.
	\end{equation}
	The necessary parameter adjustments for learning are calculated by local measurements of the equilibrium states, with the learning rule for the synaptic weights defined as follows:
	\begin{equation}
		-\frac{\partial{C}}{\partial W_{ij}} = \Delta W_{ij} \propto \frac{1}{\beta}\left[ (\rho(s_i)\rho(s_j))^{*\mathrm{nudge}} - (\rho(s_i)\rho(s_j))^{*\mathrm{free}} \right] \label{eqW_update}.
	\end{equation}
	
	Theoretical analysis and numerical experiments have demonstrated that EP progressively approximates BP in both neuronal dynamics and weight updates \cite{ernoult2019updates}. While this equivalence holds exactly under specific conditions, the two algorithms remain fundamentally distinct in their underlying principles and operational mechanisms.
	
	\section{Result}\label{sec3}
	
	\subsection{Enhancing CIM with Adam optimizer for ground state solution}
	Continuous simulated spin CIM faces amplitude heterogeneity, causing mapping errors to the discrete spin Ising model and hindering ground state search. To overcome the challenge of the CIM getting trapped in local optima, we enhance the computational performance of CIM in solving the ground state using the Adam optimizer.
	The Adam algorithm determines adaptive learning factors for each parameter by calculating the first and second moment estimates of the gradients. To minimize analog errors, we use perfectly inelastic walls at 
	$x_i = \begin{cases} 
		\operatorname{sign}(x_i) & \text{if } |x_i| > 1 \\
		x_i & \text{otherwise}
	\end{cases}$.  Additionally, we neglect the third-order terms in Eq.\ref{eqmuzhenfu}, as the inelastic walls are similar to nonlinear potential walls. The master dynamics equation for Adam-CIM is:
	\begin{equation}
		\begin{aligned}
			\dot{x}_{t-1} &= (p-1) x_{t-1} + Jx_{t-1} + h \\
			\phi_t &= \beta_1 \phi_{t-1} + (1 - \beta_1) \dot{x}_{t-1} \\
			\psi_t &= \beta_2 \psi_{t-1} + (1 - \beta_2) \dot{x}_{t-1}^2 \\
			\hat{\phi}_t &= \frac{\phi_t}{1 - \beta_1^t} \\
			\hat{\psi}_t &= \frac{\psi_t}{1 - \beta_2^t} \\
			x_{t} &= x_{t-1} - \alpha \frac{\hat{\phi}_t}{\sqrt{\hat{\psi}_t + \varepsilon}}
		\end{aligned}
		\label{eq:update}
	\end{equation}
	$v$ and $s$ represent the first and second moments of the gradient, while $\hat{v}$ and $\hat{s}$ represent the corrected versions. $\beta_1$ and $\beta_2$ are the decay factors, which are different from $\beta$ in Eq.\ref{eqW_update}. $\beta_1 = 0.90$, $\beta_2 = 0.99$, $\alpha$ is learning rate and $\varepsilon$ is a small number to prevent division by 0. The pseudo code of Adam-CIM is Algorithm \ref{Adam-CIM}.
	\begin{algorithm}[!h]
		\caption{Algorithm of Adam-CIM}
		\label{Adam-CIM}
		\begin{algorithmic}[1]
			\Require $J$,$h$,$epoch$ ,$\beta_1$,$\beta_2$,$\alpha$
			\Ensure $\mathbf{x}$
			
			\State Initialize $\mathbf{\phi}=0$, $\mathbf{\psi}=0$, $\mathbf{x}=0$, $t=1$
			\While{$t < epoch$}
			\State $\mathbf{g}_t = \left(\mathbf{p} - 1\right) \cdot \mathbf{x} + Jx + h$
			\State $\mathbf{\phi} = \beta_1 \cdot \mathbf{\phi} + (1 - \beta_1) \cdot \mathbf{g}_t$
			\State $\mathbf{\psi} = \beta_2 \cdot \mathbf{\psi} + (1 - \beta_2) \cdot \mathbf{g}_t^2$
			\State $\hat{\mathbf{\phi}} = \frac{\mathbf{\phi}}{1 - \beta_1^{t+1}}$
			\State $\hat{\mathbf{\psi}} = \frac{\mathbf{\psi}}{1 - \beta_2^{t+1}}$
			\State $x_{i} = x_{i} - \alpha \frac{\hat{\phi}_t}{\sqrt{\hat{\psi}_t + \varepsilon}}$
			\If {$\left| \mathbf{x}_i \right| > 1$}
			\State $x_i = \operatorname{sign}(x_i)$
			\Else
			\State $x_i = x_i$
			\EndIf
			\State $t \leftarrow t + 1$
			\EndWhile\\
			\Return $\mathbf{x}$
		\end{algorithmic}
	\end{algorithm}
	
	%All programs are written in Python and run on a CPU Intel i5-11600KF @ 3.90GHz and a GPU NVIDIA GeForce RTX 3060 Ti.

	%\subsection{Enhancing CIM with Adam optimizer for ground state solution}\label{sec2subsec1}
	
	To prevent the CIM from becoming trapped in local minima, we employ the Adam optimizer to enhance its performance in solving the ground state solution. 
	
\begin{figure}[h]
		\centering
		\includegraphics[width=1\textwidth]{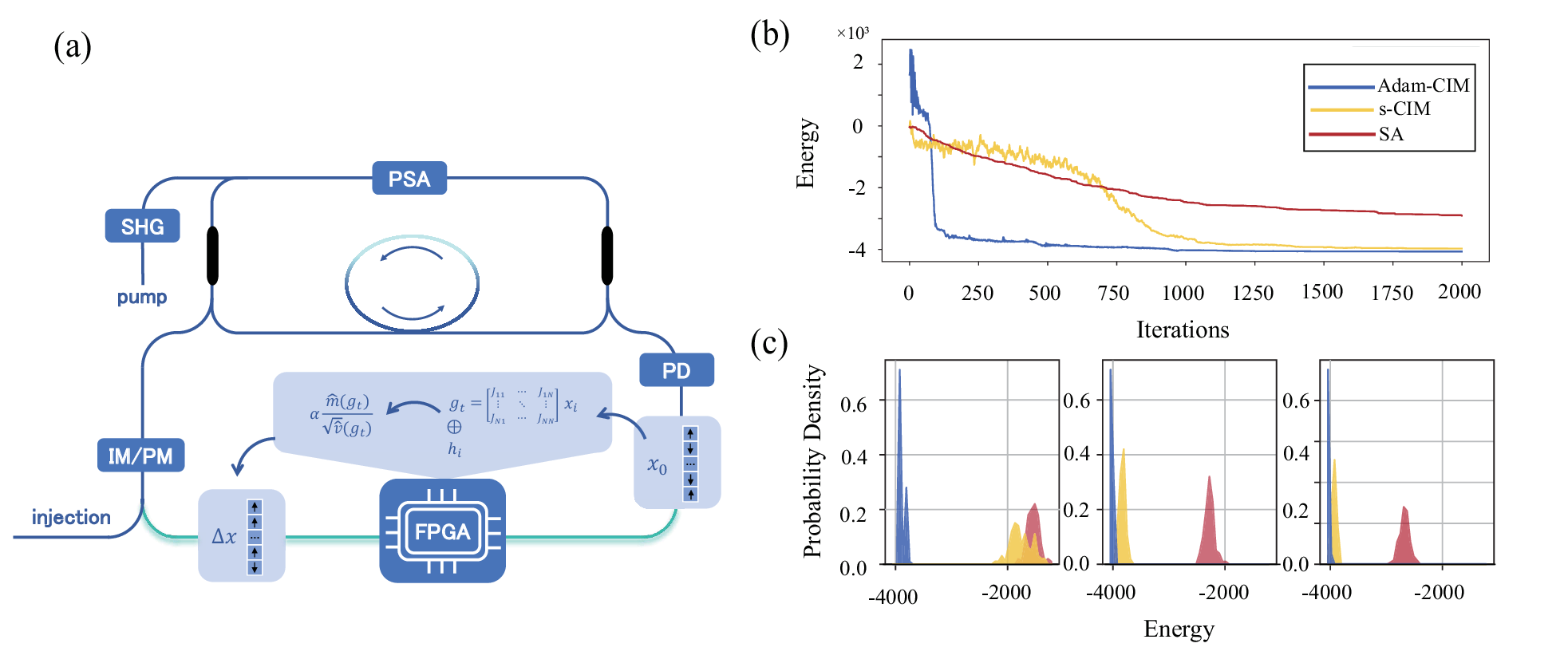}
		\caption{\textbf{Schematic diagram of the Adam-CIM and the energy distribution during evolution.} (a). Schematic diagram of the Adam-CIM. The blue loop represents the optical pulses, while the green line represents the electrical signals computed by the FPGA. The shaded area indicates its computational process, resulting in the amplitude and phase of the injected pulses. SHG: second harmonic generation, PSA: phase sensitive amplification, PD: photon detector, IM/PM: intensity/phase modulator. (b). The energy of the system during the computation of G1 for three different algorithms.  (c). The distribution of the results after 1000 calculations with the same parameters, specifically for iteration counts of 500, 1000, and 1500.} \label{fig1}
	\end{figure}

	Here, it represents the energy of a network with binary spins $\sigma_{i}$ under the constraint matrix $J_{ij}$ and the bias field $h_{i}$. Recent studies have revealed that the loss functions of certain COPs exhibit structural similarities to Ising energy formulations, inspiring novel physics-based approaches for tackling NP-hard problems. In 2013, Wang et al. proposed a CIM implemented by DOPOs and performed calculations on all instances of the Max-Cut problems \cite{wang2013coherent}. Building upon this foundation, we implement the Adam optimization algorithm to enhance the performance of CIM. As illustrated in Fig. \ref{fig1}(a), our system architecture comprises a delayed optical feedback loop integrated with an FPGA-based measurement and control system. The FPGA processes partial measurement data from the optical path to determine both amplitude and phase characteristics of the injected photons. These parameters are then precisely modulated through a phase modulator and intensity modulator before being reinjected into the fiber ring resonator. Through iterative feedback and optimization, the system progressively converges to its ground state configuration. The final spin measurement outcomes directly correspond to optimal solutions for the target of COPs. 
	
	As the pump intensity $p$ exceeds the minimum gain threshold, spontaneous symmetry breaking occurs, generating two distinct states that correspond to spin configurations. However, this mapping introduces significant amplitude heterogeneity issues that compromise the optimality of the solution. The mapped amplitudes fail to precisely represent true spin values due to their inherent physical limitations. Furthermore, the system exhibits complex dynamics where each spin evolves toward different stable points depending on its specific coupling strength $J_{ij}$. These combined effects prevent the system from reliably converging to the true ground state configuration \cite{leleu2019destabilization,lu2023combinatorial,lu2023speed,fan2025non}. 
	To address this issue, we present an Adam-optimized CIM framework for discrete variables optimization, combining the physical advantages of CIM with the computational efficiency of the Adam algorithm. This hybrid approach demonstrates superior scalability and implementation simplicity, also effective for large-scale multi-objective optimization problems. The reason is that the proposed algorithm can smooth out oscillations in the high sensitivity direction while increasing the contribution in the low sensitivity direction, thereby improving the convergence speed and helping to escape from saddle points and local minima. The detailed derivation of the Adam-CIM principle and the pseudocode can be found in the Methods section.

	In Fig.\ref{fig1}(b) and (c), we compare the computational speed and energy distribution of the three algorithms in solving the G1 graph in the Max-Cut database (G-set), respectively. Energies of the spin configuration calculated by Adam-CIM at iteration 100 drop significantly and are very close to the lowest value after 2000 iterations in Fig.\ref{fig1}(b). The s-CIM declines significantly at 700  iterations, while Simulated Annealing (SA) decreases slowly. Only Adam-CIM exhibits a brief peak in the early stages of evolution due to the first-order moment and second-order moment in the Adam algorithm being usually set to zero in initialization, resulting in unstable weight updates. Adam-CIM and s-CIM have similar final energy values, with Adam-CIM being slightly lower, while SA has the highest energy value, differing significantly from the others. Fig.\ref{fig1}(c) shows that the energy distribution of Adam-CIM is more concentrated in the low-energy parts. Obviously, the Adam-CIM algorithm outperforms the CIM and SA in both computational speed and results, achieving optimal solutions while reducing computation time. 
	
	\subsection{Fully connected neural network training with EP on Adam-CIM}\label{sec2subsec2}
	
	In this section, we implemented a fully connected neural network using the Adam-CIM and EP for recognizing handwritten digits from the MNIST database, which is a standard benchmark dataset for evaluating the performance of neural networks. The schematic diagram of training the neural network based on the Ising model is shown in  Fig. \ref{fig2}(a), with more details in the Appendix \ref{secC}.
	\begin{figure}[h]
		\centering
		\includegraphics[width=1.1\textwidth]{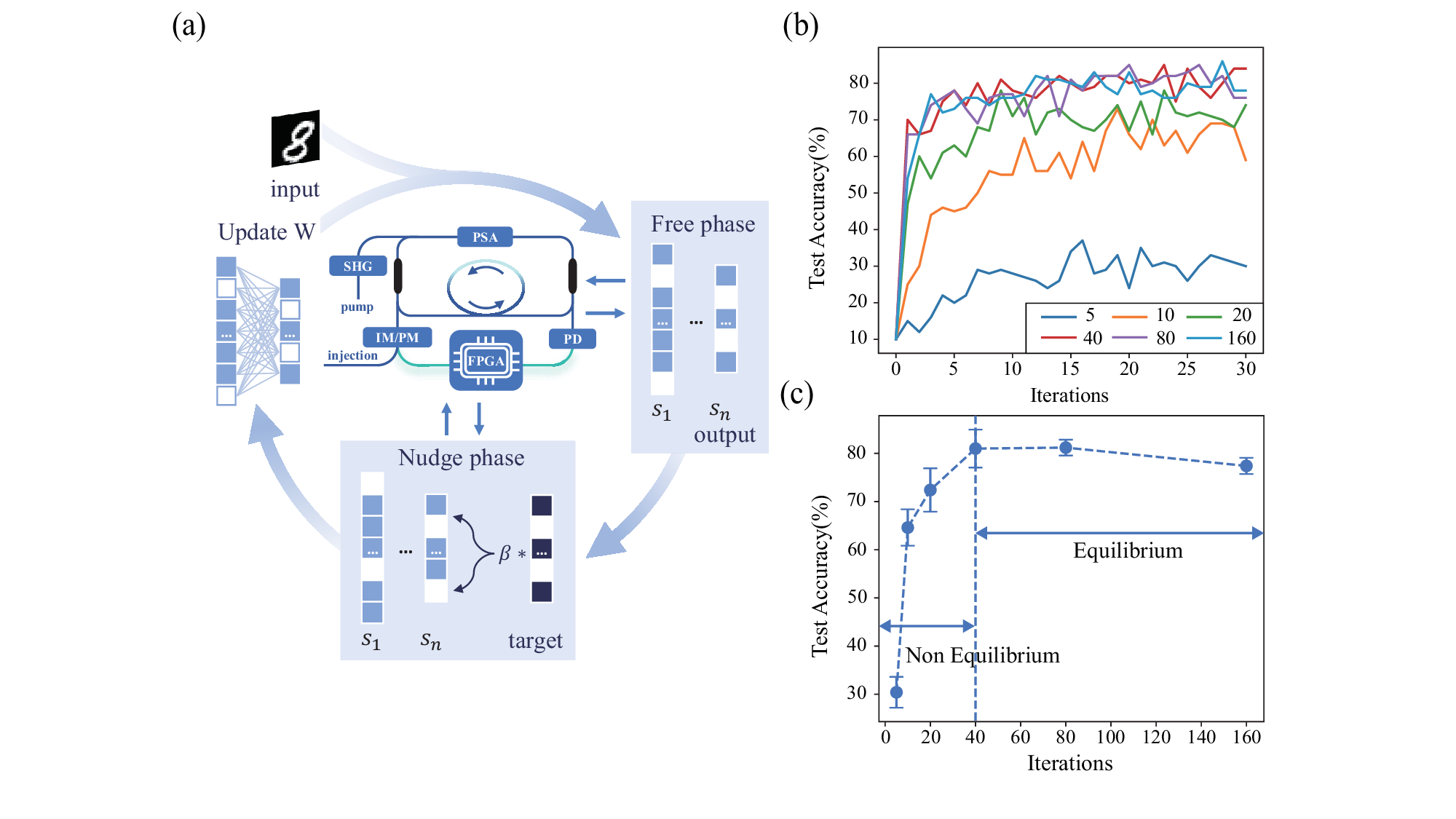}
		\caption{\textbf{Schematic diagram of training MLP by CIM and equilibrium phase transitions in training.}
			(a). Embedding the MLP architecture in the Adam-CIM with a single hidden layer consisting of 256 nodes. The input data for solving the MNIST database are \(28 \times 28\) images, where the two-dimensional image matrices are transformed into vectors. The two shaded rectangles represent the two phases in the EP, distinguished by whether the targets are used as the system's external input energy, with the value of \(\beta\) controlling it. The two phases, from the initial state to the energy local minimum, are both achieved through Adam-CIM sampling. MFB: measurement-and-feedback system. (b). Illustrates the relationship between the iterations in the Ising model during each phase and the accuracy on the test set. (c) Phase transition for the iterations of CIM when the training stabilizes.%b和c不是很确定
		}\label{fig2}
	\end{figure}
	
	While both BP and EP serve as learning algorithms for neural networks, they differ fundamentally in their operational principles and computational paradigms. BP operates through two distinct phases: Forward Propagation and Backward Propagation. The error term for a neuron in the \(l\)-th layer is the weighted sum of errors from the \((l+1)\)-th layer, multiplied by the gradient of its activation function, causing error propagation from later to earlier layers. EP requires the gradient of the loss function concerning the weights, which depends on the neuron states of all layers, meaning two Forward Propagations are needed to update the weights. Crucially, BP uses explicit analytical calculations for the gradient, while EP computes the network ground state numerically and relies on implicit functions. This makes EP biologically plausible but introduces unique computational constraints, including iterative convergence requirements, numerical approximation errors, and significant resource demands when simulating continuous physical processes on classical computers. These limitations reflect inherent challenges in digitally emulating physical systems and highlight the fundamental trade-offs between analytical precision and physically plausible computation.
	
	In addition, EP is more easily implemented using various types of Analog Hardware, whereas BP relies solely on the computational method of classical digital hardware. The training procedure is depicted in Fig.\ref{fig2}(a), and we provide a comprehensive description of how we execute the free phase, the nudge phase, and the update rule. 
	The input data for our MNIST classification task consists of $28\times28$ pixel grayscale images, where the two-dimensional image matrices are transformed into vectors. We employ an MLP architecture comprising a single hidden layer with 256 neurons.  In the following, we perform an analysis of the scalability of the number of hidden layers. The key distinction between the free phase and the nudge phase manifests through the introduction of a small perturbation term to the bias of the hidden layer. The computational process for both phases to reach equilibrium relies on Adam-CIM to find the ground state energy of the system.
	We denote the input layer as layer $0$, the state of the neurons in layer $i$ as $s_i$, the weights between layer $i$ and layer $i+1$ as $W_i$, and the bias as $W_{ibias}$.
	For continuous input, we convert them to the bias field of the $s_1$ part of the Ising model, $h_1 = W_0 \times s_0 + W_{0bias}$.
	After deforming the neural network weight matrix \(W_1\), we get the constraint matrix of Ising model $J = \left[ \begin{matrix} 0 & W_1 \\ W_1^T & 0 \end{matrix} \right]$ and the size of matrix $ J $ is  $s_1 + s_2$. 
	Upon convergence to the ground state energy configuration, the network yields stable spin configurations representing each neuron's final state.  The weight matrix \(W\) and bias \(W_{\text{bias}}\) are then updated via gradient descent by the changes of neuron states before and after the nudge. The process is repeated multiple times for updates, and the training ends when the system reaches a stable state. 
	
	Therefore, the training performance relies on the quality of the ground state solution to the Ising problem. This process begins by using CIM to identify the stable point of the ground state, followed by updating the weight matrix based on the ground state energy. The training procedure is an adiabatic process in which the weight matrix $W$ is gradually adjusted toward the target matrix $W_t$.
	The Fig.\ref{fig2}(b) illustrates the relationship between the iterations in the Ising model during each phase and the accuracy on the test set. It can be concluded that when the iterations are too low, CIM lacks sufficient time to solve the Ising model, preventing the system from reaching the ground state and resulting in a very low success rate on the test set. As the iterations increase, the success rate gradually improves. However, once the iterations exceed 40, the success rate stabilizes and no longer improves with additional iterations. Therefore, the solving performance of CIM is influenced by the solving time and is closely tied to the iterations after each weight update of $W$.
	In Fig.\ref{fig2}(c), we present the relationship between the accuracy of the test set and the iterations of CIM when the training stabilizes. When $Iterations<40$, for the given neural network parameters, Adam-CIM terminates the computation before reaching the minimum energy spin configuration. Specifically, both the free and nudge phases have not yet reached equilibrium. Updating the weights and bias parameters with the neuron states at this point leads to significant deviations, resulting in a markedly lower success rate on the test set, which improves as the iterations increase. When $Iterations>40$, the iterations are sufficient to meet the computational requirements of the neural network, the system completes the phase transition, and equilibrium is approached. The success rate on the test set stabilizes at a high level. 
	During this process, the iteration in Adam-CIM exhibits a training phase transition behavior similar to that of a ferromagnetic model when training the neural network for classification.

	\begin{figure}[h]
		\centering
		\includegraphics[width=1\textwidth]{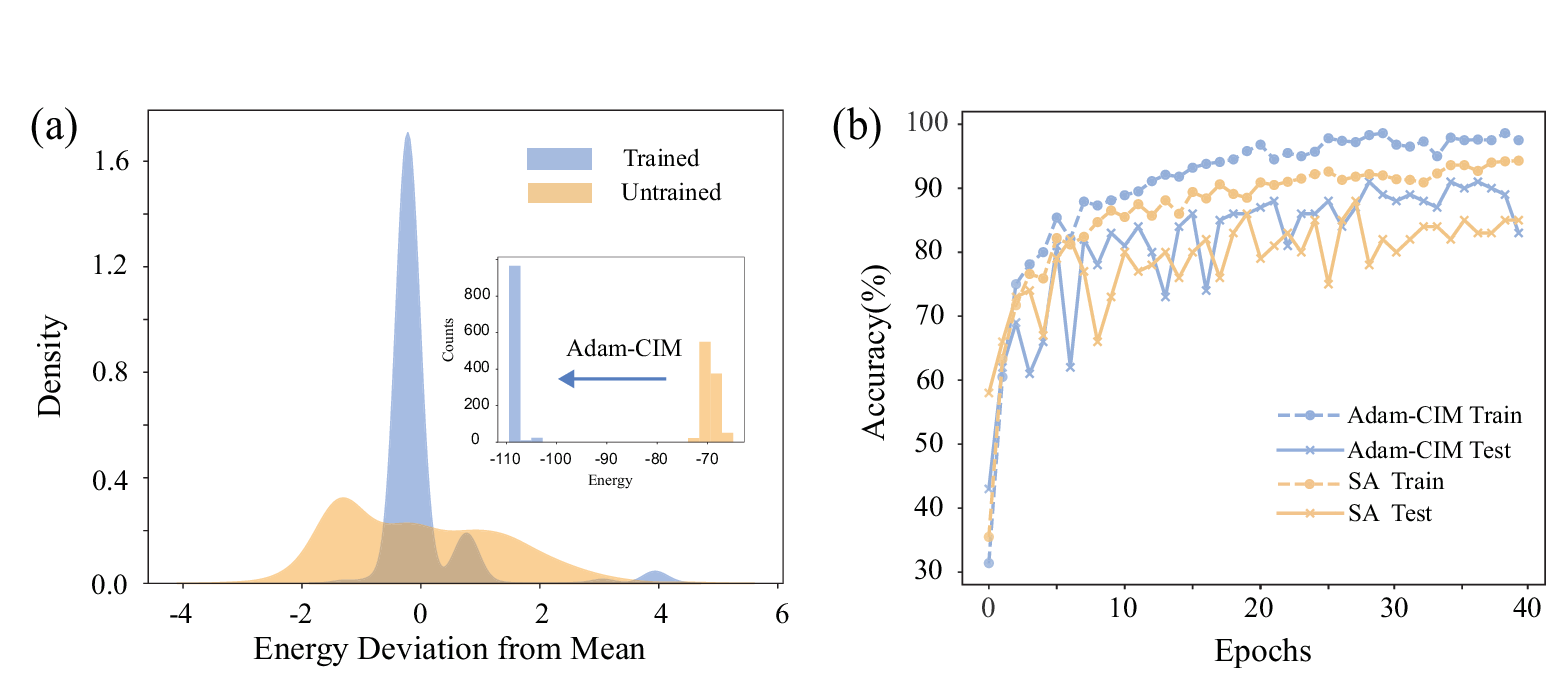}
		\caption{\textbf{Energy Distribution in Networks and Comparative Training Accuracy.}
			(a) The energy distribution of the network before and after training on 1000 images. The inset shows the frequency distribution of the energy, while the larger one uses KDE based on the former.	(b) Embedding the MLP architecture into the Adam-CIM and SA, with a single hidden layer consisting of 120 nodes.}\label{fig3}
	\end{figure}
	
	To present the training results more intuitively, we show the energy distribution before and after training in Fig.\ref{fig3}(a). At the beginning of the training, the energy distribution is relatively scattered and high. After the training, energy decreases and the distribution becomes concentrated, indicating that the training has established the specific mapping relationship between the input and the corresponding output. 
	Kernel Density Estimation (KDE) was also used to display the distribution of energy through a smoothed peak curve. 
	The probability distribution becomes denser, corresponding to a decrease in entropy, as the system transforms from a disordered distribution to an ordered one, resulting in a decrease in free energy. This process is similar to the magnetic phase transition that occurs in ferromagnetic systems when the temperature decreases.
	Another important indicator is the test accuracy after training. Here we compare Adam-CIM with SA in Fig.\ref{fig3}(b), which benchmarks the results obtained on the ref.\cite{laydevant2024training}.
	Initially, the annealing temperature is high, enabling the system to explore a wide range of spin configurations. Subsequently, the temperature is gradually reduced and the system becomes steady at the end of the annealing process (details see Methods). Adam-CIM outperforms SA on both the training and test sets, as SA is more prone to getting stuck in local optima when solving ground states.
	\begin{figure}[htbp]
		\centering
		\includegraphics[width=1\textwidth]{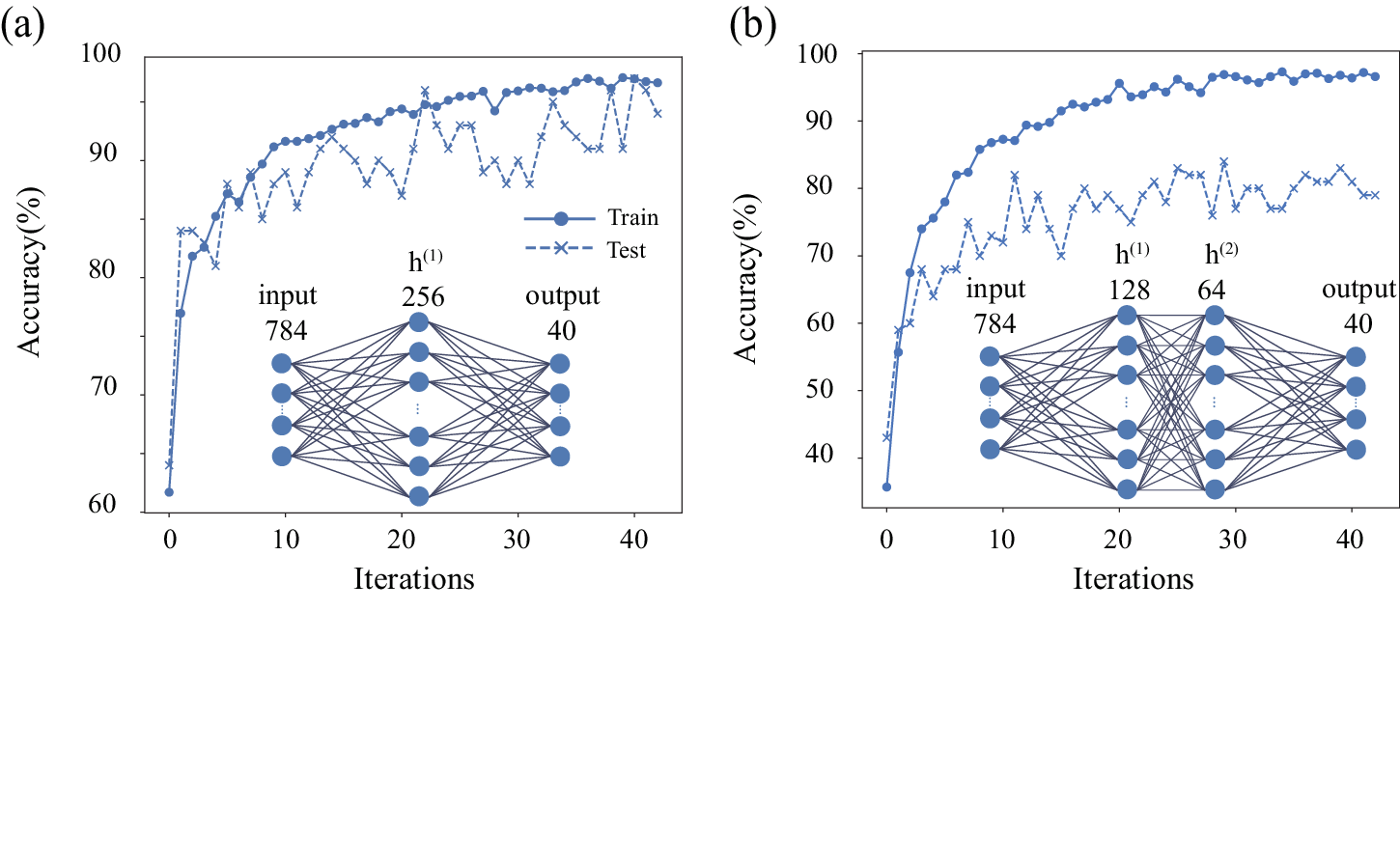}
		\caption{\textbf{Train MLP with different structures and accuracy.}(a) and (b) show the MLP structure with one and two hidden layers, respectively, along with their corresponding accuracy changes as the iterations progress.}\label{fig4}
	\end{figure}
	We trained an MLP with one hidden layer based on CIM as shown in Fig.\ref{fig4}, and its classification accuracy on the test set can reach $95\%$.
	Here, although we need to classify images into ten categories for the recognition of the MNIST database, the number of output layers here is 40 instead of 10. Each category corresponds to four neurons, and the final recognition result is the average of the outputs of the four neurons, which can avoid the impact of extreme data on the recognition performance.  
	The choice of network structure plays a significant role in the training outcomes when utilizing CIM for neural network training. Next, we will investigate various network architectures and their scalability during the training process.
	
	\subsection{Scalability analysis of neural network structure}\label{sec2subsec3} 
	
\begin{figure}[h]
		\centering
		\includegraphics[width=1.0\textwidth]{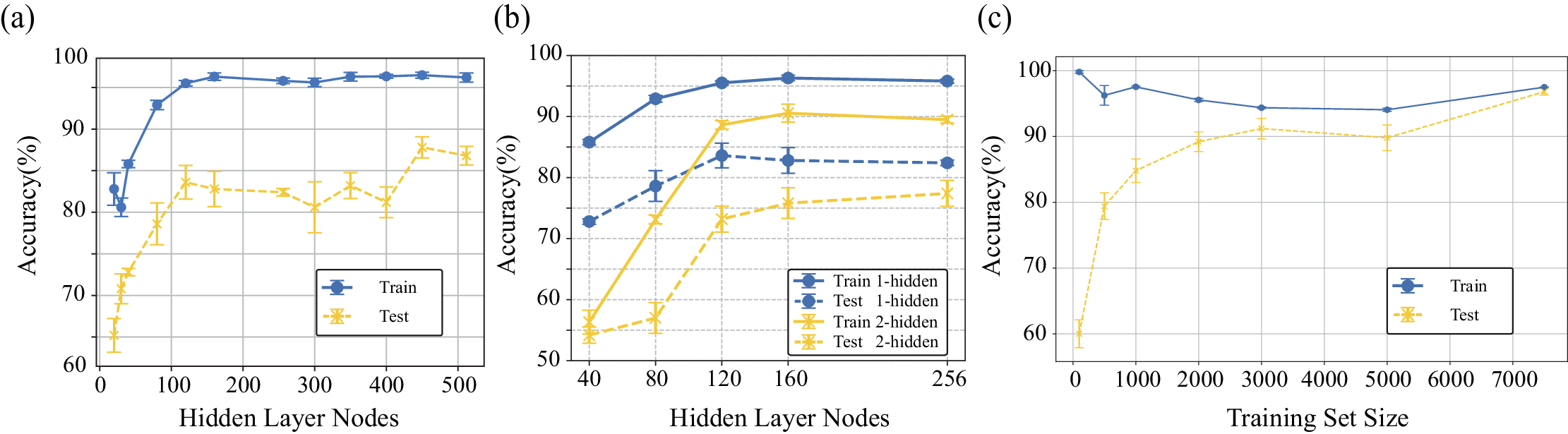}
		\caption{\textbf{Scalability study of CIM-based training MLP}(a). Effect of the number of nodes on accuracy, where the number of hidden layers is one. (b). In the case with two hidden layers, the number of nodes in each layer is half of the corresponding x-axis value, ensuring that the total number of parameters across the entire network remains consistent. (c). Training and testing accuracy as functions of training set size.}\label{fig5}
	\end{figure}

	In this section, we study the structural scalability of neural networks trained using CIM, focusing on three key architectural dimensions: the number of nodes, the number of network layers, and the network architecture.
	The selection of appropriate hyperparameters, such as the number of hidden layer nodes and the training set size, plays a crucial role in determining the performance of neural networks for both regression and classification tasks. While a deeper network with more nodes theoretically enhances its ability to fit the data, this can also lead to overfitting, increased training complexity and duration, difficulties in model convergence, and higher computational resource consumption.
	To systematically investigate these relationships, we conducted comprehensive architectural experiments by varying the structure and number of neurons in the hidden layers. 
	Fig.\ref{fig5}(a) shows the effect of the number of hidden layer nodes on accuracy, where the number of hidden layers is set to 1. The accuracy increases initially at $80\%$ for small nodes and then stabilizes at larger than $95\%$ for more than 100 nodes on the training set. When the nodes are small, the network is underfitting, meaning that the network parameters cannot fully extract recognizable features. Once the nodes reach the threshold, it indicates the training set is insufficient to train all neurons, and accuracy will not be greatly improved. 
	
	Next, we increased the number of hidden layers in Fig.\ref{fig5}(b). With the same number of nodes, the train and test accuracy of 2-hidden layers is lower than that of 1-hidden layer. Theoretically, augmenting network depth through additional hidden layers can significantly enhance learning capacity by progressively transforming the feature representations extracted by preceding layers into higher-level abstractions. However, this architectural complexity necessitates more sophisticated training algorithms to ensure effective parameter optimization, particularly in deep network configurations. Our findings suggest promising directions for future research in multi-layer architectures, which is also related to the theory of EP. A similar conclusion can be found in \cite{wang2402training}, which suggests that a complex network does not necessarily improve accuracy certainly but can even cause a slight decrease. We found that as the network becomes complex, a larger $\beta$ is often required, while this is still insufficient to outperform simpler architectures. As can be seen from Eq.\ref{eqW_update}, EP updates the weight $W$ based on the neuron states after evolution through the free and nudge phase at equilibrium. 
	As the network complexity increases, accurately determining the spin configuration of the ground state energy becomes more challenging. Errors are amplified during implicit transmission, making EP more suitable for simpler architectures 
	Furthermore, it should be emphasized that our implementation employs classical numerical simulation of the Adam-CIM dynamics, which introduces additional approximation errors compared to actual physical implementations
	
\begin{figure}[h]
		\centering
		\includegraphics[width=1.0 \textwidth]{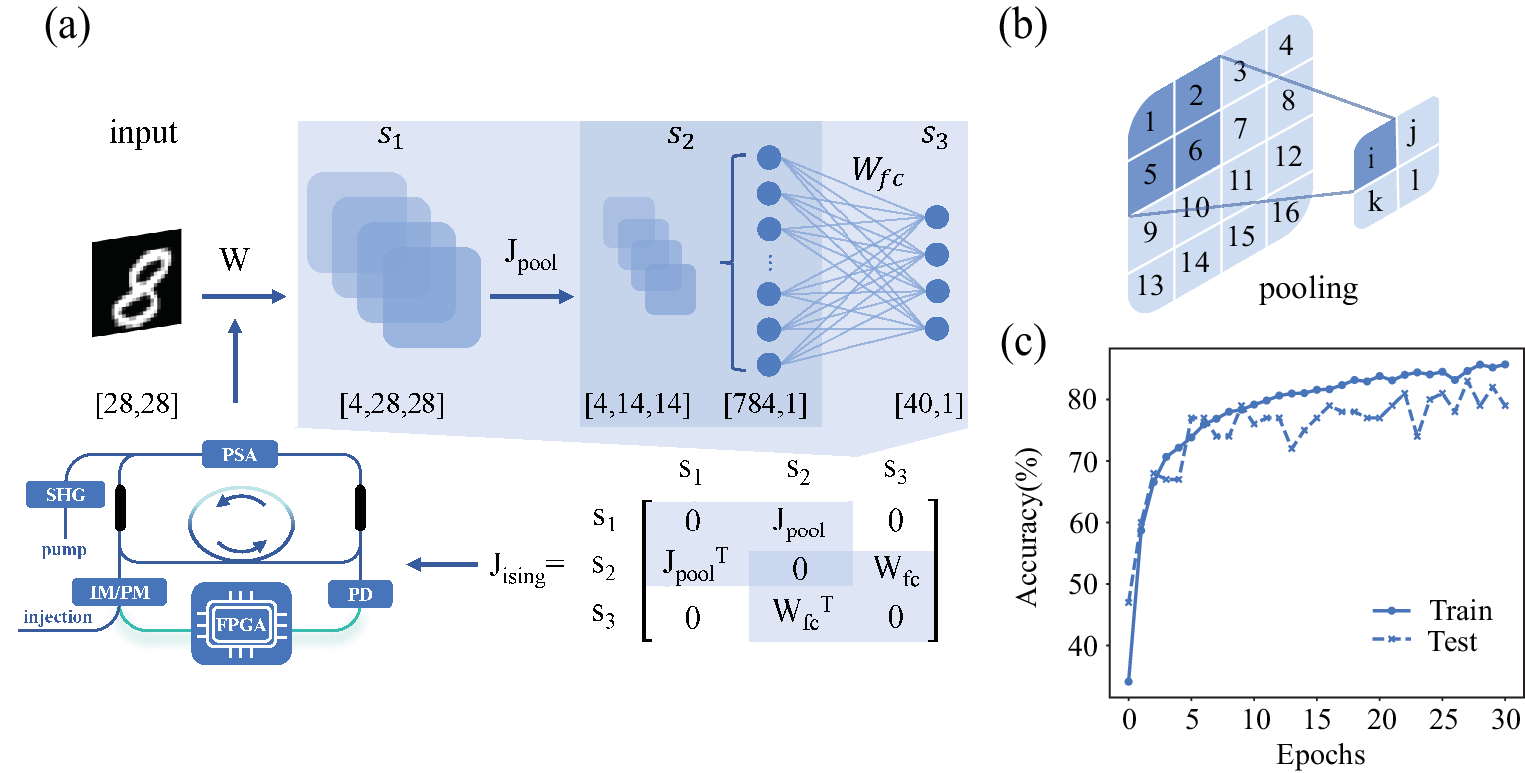}
		\caption{\textbf{Schematic diagram of training CNN based on CIM}(a). The structure of a single-layer convolutional neural network implemented by Adam-CIM, which includes four different convolution kernels. \( J_{pool} \) is the weight matrix between \( s_1 \) and \( s_2 \) (with \( W_{fc} \) being the weight matrix between \( s_2 \) and \( s_3 \)), and the dimensions of \( s_1 \), \( s_2 \), and \( s_3 \) are annotated below each of them. Adam-CIM samples based on \( J_{ising} \) to obtain the ground state spin configuration. (b). The convolution process implemented by \( J_{pool} \). (c). The evolution of accuracy concerning iterations.
		}\label{fig6}
	\end{figure}

	Building upon our MLP results, we further demonstrate the versatility of our framework by extending CIM-compatible network structures to convolutional neural networks (CNNs). Specifically, we implement EP with Adam-CIM optimization to train a CNN on the MNIST dataset.  CNNs are particularly well-suited for image processing tasks due to their hierarchical feature extraction mechanism, which traditionally comprises four key components: a convolutional layer, a pooling layer, a fully connected layer, and an output layer. As shown in Fig.\ref{fig6}(a), features can be obtained by sliding different filters on the input data. Then Adam-CIM samples the pool and fully connected layers as a whole to reach the local minimum energy state, and uses the obtained neuron state to update the weights according to the EP. In Fig.\ref{fig6}(c), test accuracy is around $80\%$ after 20 epochs. The result still has some gap compared to the improved CNN \cite{bhatt2021cnn}, due to the structure of a CNN being more complex than MLP, which increases the difficulty of finding the ground state spins, and the increase in layers makes it difficult for gradients to propagate. These effects collectively limit the use of inaccurate neuron states to update the weights, deviating from the correct gradient descent direction, resulting in a decrease in accuracy. The complexity of the network does not necessarily lead to an effective improvement when using EP. In summary, the CNN implemented with the Adam-CIM and EP can essentially achieve recognition on the MNIST, and with optimization, it can have broader application scenarios.
	
\begin{figure}[h]
		\centering
		\includegraphics[width=0.7 \textwidth]{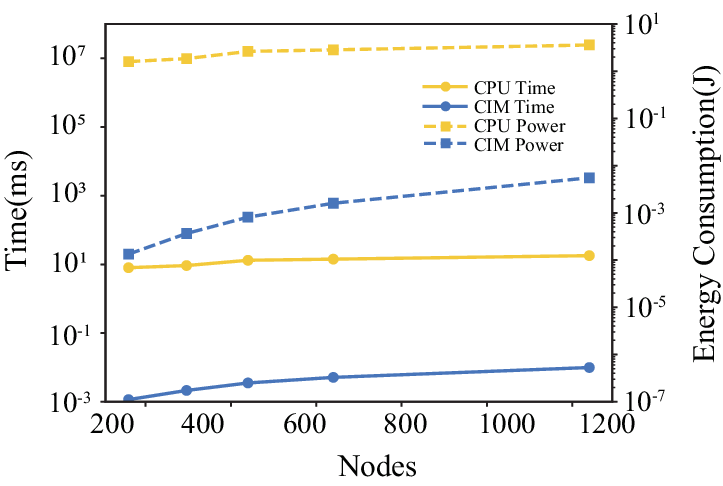}
		\caption{\textbf{Estimated Time and Energy Consumption for Training Networks with Optical CIM and CPU.} 
			The dashed and solid lines represent the variations of time and energy consumption concerning the nodes. The yellow line represents the CPU, and the blue line represents the CIM. }\label{fig7}
	\end{figure}

	Furthermore, photonic Ising machines and integrated optics have experienced rapid development over the past decade.  Assuming we use CIM with the 100 GHz optical frequency comb integrated chip proposed in \cite{reifenstein2021coherent}, with a signal source soliton frequency comb generator consuming 100mW and an EOM modulator consuming 400mW \cite{stern2018battery,wang2018integrated}, we can achieve approximately three orders of magnitude in both time and energy efficiency advantages in Fig.\ref{fig7}.
	
	\section{DISCUSSION}\label{sec13}
	
	Optical computing has emerged as a promising paradigm for next-generation computing, particularly in quantum information processing and artificial intelligence applications, owing to its inherent advantages in ultra-high speed (potentially $>100 GHz$) and ultra-low energy consumption ($<1 fJ/operation$). However, conventional neural network training methodologies predominantly rely on BP with gradient descent optimization. The BP typically is easy to implement in electronic computing but faces significant implementation challenges in optical domains due to the difficulty of performing precise chain rule differentiations in photonic hardware \cite{hughes2018training,pai2023experimentally}. In contrast, EP is a biologically inspired learning algorithm, theoretically introduced in the context of leaky-integrate neurons \cite{scellier2017equilibrium}. It conforms to the evolution rules of physical systems and has been proven to achieve results close to BP under certain conditions. Therefore, this makes EP particularly suitable for training photonic neural networks, bridging the gap between biological plausibility, physical realizability, and computational efficiency.
	
	%%%%%%%%%%%%%%%
	EP requires a physical system to reach equilibrium under two different conditions to update weights. Our hybrid computing platform combines this paradigm with CIM based on DOPOs can spontaneously evolve to the spin configuration of the ground state energy and is scalable. In this work, we studied the improved Adam-CIM algorithm, and the simulation results show that it evolves faster and gets lower energy than traditional CIM and SA. We compared related works on training neural networks using CIMs in Tab.\ref{tab1}. Compared to RBM, EP has a higher accuracy upper limit, and it can be further accelerated with a GPU. So it reduces training time because as the network structure becomes complex, the training time increases significantly, which is difficult to accept in practical applications. By porting Adam-CIM to GPU, the computation time can be reduced by about one order of magnitude, with no significant change in accuracy. Meanwhile, the Adam-CIM can also improve the training performance. For example, with 120 hidden-layer nodes, the test accuracy of Adam-CIM is higher than that of SA simulation and D-Wave. In addition, we found that EP can achieve great recognition performance with simpler structures. Increasing complexity does not lead to a significant improvement in accuracy but increases training time and requires a larger $\beta$ value. This is the most important hyperparameter in EP. The training heavily depends on $\beta$, so selecting an appropriate value increases the total workload. This is also a direction where EP can be optimized in the future. By optimizing the structure, it can be made more robust. A certain range of $\beta$ values or dynamically updating $\beta$ based on partial measurement results can achieve the same training effect as the optimal $\beta$ is required. Last but not least, Adam-CIM has a significant advantage in terms of scalability compared to its implementation on D-Wave \cite{laydevant2024training}. The latter can only support a maximum of 120 neurons in the hidden layer. Additionally, the number of images used for training the network is also limited to 1000. Currently, the maximum spin of CIM is 100,000 for now, far surpassing 5,000 spins of D-Wave. It can use more hidden layer neurons and realize more complex CNNs. This provides development in applying this technology to fields such as computer vision and image processing. The structure of EP can be optimized to reduce the gap with BP when updating gradients. For example, introducing three-phase \cite{laborieux2021scaling} removes bias and brings EP performance on CIFAR-10 closer to that achieved by BP. One important consideration is that, as an implicit update algorithm without a defined gradient formula, the update effect of EP can only be continuously optimized to approach BP, but cannot surpass BP. This is a rather serious limitation for the future development of EP.
	
	\begin{table}[h]
		\centering
		\caption{Performance comparison of different Ising machine implementations on MNIST database classification}\label{tab:ising_comparison}
		\begin{tabularx}{\textwidth}{p{1cm} X p{2cm} p{1.2cm} @{\hspace{1cm}} p{2cm}}
			\toprule
			& \textbf{Operation principle} & \textbf{Database} & \textbf{Accuracy} & \textbf{Network structure} \\
			\midrule
			This work & EP on Adam-CIM & MNIST database & 96.8\% ($\pm$ 0.52\%) & 1-fc (256 hidden units) \\
			\addlinespace
			\cite{laydevant2024training} & EP on Quantum annealing (D-Wave) & MNIST database & 88.8\% ($\pm$ 1.5\%) & 1-fc (120 hidden units) \\
			\addlinespace
			\cite{bohm2022noise} & \raggedright Opto-electronic Ising machine with injecting noise & 8$\times$8 greyscale digits & 94.3\% & RBM (100 hidden, 64 visible) \\
			\addlinespace
			\cite{niazi2024training} & Sparse Ising Machine & MNIST database & 92\% & RBM (4096 hidden units) \\
			\addlinespace
			\cite{song2023training} & QUBO on Ising Machine & Downsampled to 2$\times$2 (only digits 6,9) & 98.3\% & QNN (1 hidden layer) \\
			\addlinespace
			\cite{PhysRevLett.131.063801} & \raggedright Spatial Photonic Ising Machine & MNIST with low-rank interactions & $\sim$90\% & RBM (no hidden units) \\
			\addlinespace
			\cite{ulanov2019quantum} & SimCIM for Boltzmann sampling & MNIST database & 86.9\% & Fully-connected BM \\
			\bottomrule
		\end{tabularx}\label{tab1}
		\footnotetext{fc: fully-connected layer, RBM: Restricted Boltzmann Machine, QNN: Quantum Neural Network, BM: Boltzmann Machine.}
	\end{table}
	
	\section{Appendix}\label{sec11}
	
	%\textbf{CIM numerical simulation method}\label{secA}
	\textbf{Simulated Annealing Algorithm}\label{secB}
	
	SA is a general probabilistic algorithm that can randomly update states and search for the optimal solution to COPs. It compares the problem-solving process to the annealing process of solid, and uses the temperature-dependent escape probability to perform a global search in the solution space. 
	
	We use the logarithmic form to decrease the temperature $T_i = T_0 \ln\left(1 + \frac{i}{n_{\text{iter}}}\right)$, where $n_{\text{iter}}$ represents the number of iterations in each phase, and $T_0$ is the initial temperature. This method is smoother than traditional linear or exponential decay methods, preventing the temperature from decreasing too quickly. It has been used in many experiments with good results. During the update process, the energy difference $\Delta E = s_{i, \text{flipped}} \left(\sum_i J_{ij} s_j + h_i\right)$ may occur due to spin flipping. Finally, we update the spin probability according to the Metropolis Acceptance Criterion, where the spin flip probability $P \propto \exp\left(-\frac{\Delta E}{T}\right)$.
	
	We modified and implemented the simulated annealing sampler based on the D-Wave algorithm package. Each sampling is performed multiple times to obtain a reliable ground state estimate. This algorithm is implemented on the Python platform and can be found in our CIM training neural network code.
	
	\textbf{Mapping neural network structures to CIM}\label{secC}
	
	Although its structure has similarities to the Ising model, it cannot be directly mapped to CIM. First, the nodes and data are often continuous, whereas in CIM, they are binary spins of $\pm 1$. Second, how the Ising machine updates the constraint matrix by sampling the ground state for a fixed constraint matrix.
	
	We simplify the neurons to binary values, and add the product of the input layer data and the weight matrix as the bias of the hidden layer.
	We denote by $s$ the state variable of the network, $v$ the target of data set, and $\theta$ the set of parameters to be learned. The neurons should be simplified to binary values, and the product of the input layer data and the weight matrix is added as the bias of the hidden layer. The update of the network gradient can be performed by EP. Considering the fixed $v$ and $\theta$, in the free phase of EP, fixed point $s_{\theta,v}^0$ satisfies $\frac{\partial F}{\partial s}(\theta,v,0,s_{\theta,v}^0) = 0,$ (which corresponds to the nudge phase satisfying $\frac{\partial F}{\partial s}(\theta,v,\beta,s_{\theta,v}^\beta) = 0$ corresponding a local minimum of $F$).
	%这能量不用E吗？
	
	The CIM can be used for rapid sampling to obtain the stable state \( s_{\theta,v}^0(s_{\theta,v}^\beta) \). Under the constraint matrix \( J \), the minimum gain principle is used to evolve the system to the ground state. The system energy satisfies:
	\begin{equation}
		V(s) = \frac{1 - p(t)}{2} s_i^2 - \frac{1}{2} \sum_{i,j} J_{ij} s_i \, \text{sign}(s_j) - \sum_{i} h_i s_i, \quad \left| s_i \right| \le 1 \ \text{otherwise} \ V = \infty,\label{eqsystem_energy}
	\end{equation}
	Here, we consider replacing the nonlinear barrier with the sign function, as this can accelerate the computation \cite{goto2021high}. Now, the corresponding state $s$ satisfies the dynamical evolution:
	
	\begin{equation}
		\frac{ds_i}{dt} = -\frac{dV}{ds_i} = [p(t) - 1] s_i + \sum_{i,j} J_{ij} \, \text{sign}(s_j) + \sum_{i} h_i.\label{eqevolution}
	\end{equation}
	
	When \( p(t) \) is less than the dissipation term \( -s_i \), the evolution direction of \( s_i \) depends on the last two terms on the right side of Eq.\ref{eqevolution}. Therefore, the system evolves in the direction of the Ising energy difference \( \Delta E = \Delta s_i \left[ \sum_{i,j} J_{ij} \, \text{sign}(s_j) + \sum_{i} h_i \right] < 0 \), leading to the ground state spin configuration of system.
	
	We define the objective function as: $F = \sum_{i,j} J_{ij} s_i s_j + \sum_{i} h_i s_i + \beta \sum_{i} (s_i - v_i)^2$, and he loss function is then given by: $C = \frac{1}{2} \sum_i \| s_i^0 - v \|^2 = \frac{\partial F}{\partial \beta} ( \theta, v, 0, s_{\theta, v}^0 )$. At the same time, $\left[ \frac{d}{d\theta} \frac{\partial F}{\partial \beta} \left( \theta, {v}, \beta, s_{\theta, v}^{\beta} \right) \right]^T = \frac{d}{d\beta} \frac{\partial F}{\partial \theta} \left( \theta, {v}, \beta, s_{\theta, v}^{\beta} \right)$, for small \( \beta \), this can be approximated as$
	\Delta \theta = - \frac{1}{\beta} \left[ \frac{\partial F}{\partial \theta} \left( \theta, v, \beta, s_{\theta, v}^{\beta} \right) - \frac{\partial F}{\partial \theta} \left( \theta, v, 0, s_{\theta, v}^{0} \right) \right]$ By CIM sampling, we obtain the free phase state \( s_i^0 \) and the nudge phase state \( s_i^\beta \). The updates for the weight matrix and bias are as follows:
	\begin{align}
		-\frac{\partial C}{\partial W_{ij}} &= \Delta W_{ij} = -\frac{1}{\beta} \left( s_i^\beta s_j^\beta - s_i^0 s_j^0 \right) \\
		-\frac{\partial C}{\partial h_i} &= \Delta h_i = -\frac{1}{\beta} \left( s_i^\beta - s_i^0 \right).
	\end{align}
	Finally, we can update the weights and biases according to the gradients:
	\begin{equation}
		W(h) \gets W(h) + \eta \Delta W(\Delta h).\label{equpdate_weights}
	\end{equation}
	Here, \(\eta\) is the learning rate.
	
	For the CNN, it can be viewed as the front part of a fully connected neural network classifier, with convolutional layers and pooling layers added for feature extraction. For the convolutional layers, due to their unique processing method, the gradient update method of the fully connected network is no longer applicable. Fortunately, there are related studies combining EP with CNNs\cite{laydevant2021training}, and the gradient update rule satisfies:
	\begin{equation}
		\Delta W_{i,i+1} = -\frac{1}{\beta} \left( s_i^{\beta} * sup_{i+1}^{\beta} - s_i^{0} * sup_{i+1}^{0} \right).\label{equpdate_weight}
	\end{equation}
	
	Here, \(*\) is convolution operation, and $ {sup}_{i+1} $ denotes the result of the unpooling of the neuron \( s_{i+1}^\beta \) in the $(i+1)$-th layer. To facilitate the subsequent pooling operations mapped into the CIM, we choose average pooling. To implement the pooling matrix, we set the matrix elements of connected neurons before and after the pooling layer to 1, as shown in Fig.\ref{fig6}(b). The positive interactions ensure that the spins of pooled neurons are as consistent as possible with the neurons in the previous layer.
	
	\backmatter
	
	\bmhead{Acknowledgements}
	The authors gratefully acknowledge the support from the National Natural Science Foundation of China through Grants Nos. 62461160263,  62401628, and 62371050.
	
	\bmhead{Data availability}
	The G-set instances used for the tests are available at https://web.stanford.edu/$\sim$yyye/yyye/Gset/. The authors declare that all data required to evaluate the conclusions of this paper are included within the paper. Additional data can be obtained from the corresponding author upon reasonable request.
	
	%\bmhead{Code availability}
	%The code to reproduce the results is available on GitHub at the following link: XXXXXXXXX
	
	\section*{Declarations}
	The authors declare no conflicts of interest.
	\section*{Author contribution}
	C.W. conceived the research project. C.F. and B.L. constructed the scheme. C.F. built the theoretical model with assistance from B.L., J. W and C.W. C.F., Z.Z., R.Z., B.L carried out the experiment, C.F., J.W and B.L. performed data processing. C.F., B.L. and C.W. wrote the manuscript. All authors have read and approved the final version of the manuscript.
	
	%\begin{itemize}
	%\item This work is funded by the National Natural Science Foundation of China (Grants No.62401628,62371050).
	%\item Conflict of interest/Competing interests (check journal-specific guidelines for which heading to use)
	%\item Ethics approval and consent to participate
	%\item Consent for publication
	%\item Materials availability
	%\item Data and Code availability :We utilised a public dataset and will publish our code at the conclusion of the review process. Simultaneously, we provide pseudo-algorithms in the method.
	%\item Author contribution
	%\end{itemize}

	%%===========================================================================================%%
	%% If you are submitting to one of the Nature Portfolio journals, using the eJP submission   %%
	%% system, please include the references within the manuscript file itself. You may do this  %%
	%% by copying the reference list from your .bbl file, paste it into the main manuscript .tex %%
	%% file, and delete the associated \verb+\bibliography+ commands.                            %%
	%%===========================================================================================%%
	
	\bibliography{sn-bibliography}% common bib file
	%% if required, the content of .bbl file can be included here once bbl is generated
	%%\input sn-article.bbl
	
\end{document}